\documentclass[11pt]{article}
\usepackage{eamt26}
\usepackage{amsmath}
\usepackage{array}
\usepackage{times}
\usepackage{url}
\usepackage{latexsym}
\usepackage[small,bf]{caption} 
\usepackage{listings}
\nocopyrightfn

\title{To Write or to Automate Linguistic Prompts, That Is the Question}
  \author{Marina Sánchez-Torrón \hspace{2em} Daria Akselrod\textsuperscript{*} \hspace{2em} Jason Rauchwerk\textsuperscript{*} \\
  Smartling \\ 244 Fifth Avenue Suite 1471 New York, NY 10001 \\
  \texttt{\{msancheztorron,dakselrod,jrauchwerk\}@smartling.com}\\}

\begin{document} 
\maketitle
\renewcommand{\thefootnote}{\fnsymbol{footnote}}
\footnotetext[1]{Equal contribution.}
\begin{abstract}
LLM performance is highly sensitive to prompt design, yet whether automatic prompt optimization can replace expert prompt engineering in linguistic tasks remains unexplored. We present the first systematic comparison of hand-crafted zero-shot expert prompts, base DSPy signatures, and GEPA-optimized DSPy signatures across terminology insertion, translation and language quality assessment, evaluating five model configurations. 
Results are task-dependent. For terminology insertion and translation, GEPA-optimized prompts are competitive with expert prompts: most differences are not statistically significant, and optimization significantly improves glossary term match rates for several models. In language quality assessment, expert prompts achieve stronger error detection while optimization improves characterization.  Across tasks, GEPA elevates minimal DSPy signatures, often closing the gap to expert performance. We note that the comparison is asymmetric: GEPA optimization searches programmatically over gold-standard splits, whereas expert prompts require in principle no labeled data, relying instead on domain expertise and iterative refinement.

\end{abstract}

\section{Introduction}
LLMs have shown remarkable capabilities across a wide range of NLP tasks. However, research shows that LLM performance is highly sensitive to the way prompts are worded and structured. Sclar et al.~\shortcite {sclar2024quantifyinglanguagemodelssensitivity} found that open-source LLMs exhibit significant performance differences when tested under slight, meaning-preserving variations in prompt formatting, such as adding or omitting punctuation marks; Voronov et al.~\shortcite{voronov2024mindformatconsistentevaluation} found that variations in how examples are selected, verbalized or separated  in few-shot settings results in performance differences, while Mizrahi et al.~\shortcite{mizrahi-etal-2024-state} found that both manual and automated prompt paraphrasing lead to drastic performance variations in both open-source and closed models. While metrics have been proposed to evaluate this fragility~\cite{errica-etal-2025-wrong,sclar2024quantifyinglanguagemodelssensitivity}, improving prompts in response to such diagnoses remains a largely manual, iterative process requiring task-specific knowledge. This practice, referred to as prompt engineering, is the subject of a substantial body of research, producing for example catalogs of prompt patterns \cite{white2023promptpatterncatalogenhance} and prompting techniques \cite{sahoo2025systematicsurveypromptengineering}.

At scale, prompt engineering presents practical challenges. First, the process is inherently trial-and-error: practitioners iteratively test and revise prompts with no systematic way to determine whether their current prompt is optimal. Second, effective prompts do not necessarily transfer across setups: a prompt tuned for one model, task, or language pair may underperform when any of these changes, and model updates can degrade previously effective prompts. 

These limitations have motivated research on automatic prompt optimization, progressing from continuous gradient-based methods to discrete approaches (see Section~2). Recent gradient-free discrete optimizers~\cite{yang2024largelanguagemodelsoptimizers,guo2025evopromptconnectingllmsevolutionary,agrawal2025gepareflectivepromptevolution} are directly applicable to the black-box API models on which LSPs typically rely, yet whether they can match expert-crafted prompts on specialized linguistic tasks remains unexplored.

We investigate this question across three linguistic tasks where language service providers (LSPs) increasingly rely on LLMs and where prompt design directly affects production quality: (a) terminology insertion, where an existing translation is corrected to conform to a provided glossary; (b) translation; and (c) language quality assessment (LQA), where translation errors are detected and characterized using an MQM-compliant \cite{Lommel2014} schema. 

This paper contributes to the current body of research by providing, to the best of our knowledge, the first comparison of expert-crafted prompts vs. optimized prompts for specialized linguistic tasks, with signatures and results available on GitHub \footnote{
\url{https://github.com/msancheztorron-smartling/EAMT2026submission}}

\section{Related Work}

\paragraph{Automatic Prompt Optimization.} The well-documented fragility of LLM performance under prompt variation~\cite{sclar2024quantifyinglanguagemodelssensitivity,voronov2024mindformatconsistentevaluation,mizrahi-etal-2024-state} has motivated a growing body of work on automatic prompt optimization. Early approaches treated the prompt as a learnable vector in the model's embedding space, optimizing it via gradient-based~\cite{liang-2021-prefix,liu-etal-2022-p} or reinforcement learning~\cite{zhang2022temperatesttimepromptingreinforcement} techniques. These methods require access to model gradients, ruling out black-box API models, and produce continuous representations that are not interpretable as natural language. Discrete approaches operate directly over natural language tokens, producing human-readable prompts. Earlier discrete methods still relied on gradient access~\cite{shin-etal-2020-autoprompt,shi2022humanreadableprompttuning,deng-etal-2022-rlprompt,zhang2022differentiablepromptmakespretrained,pryzant2023automaticpromptoptimizationgradient}, but recent work has shown that effective discrete optimization is possible without access to model parameters~\cite{yang2024largelanguagemodelsoptimizers,guo2025evopromptconnectingllmsevolutionary,agrawal2025gepareflectivepromptevolution}, making automatic optimization directly applicable to the black-box API models on which LSPs typically rely.

Despite these advances, research directly comparing automatically optimized prompts against expert-crafted prompts remains limited. Existing studies demonstrate the benefits of prompt optimization but typically benchmark against generic baselines rather than domain-expert prompts, and focus on coding, classification, and question-answering rather than specialized linguistic tasks. For example, Agrawal et al.~\shortcite{agrawal2025gepareflectivepromptevolution} show that their GEPA optimizer outperforms Group Relative Policy Optimization (GRPO) and other optimizers, while Yang et al.~\shortcite{yang2024largelanguagemodelsoptimizers} and Yuksekgonul et al.~\shortcite{yuksekgonul2024textgradautomaticdifferentiationtext} show improvements over human-designed prompts without reference to the authors' expertise. Conversely, Zhou et al.~\shortcite{zhou2023revisitingautomatedpromptingactually} found that simple manual prompts outperformed gradient-based automated prompting in the majority of setups, and He et al.~\shortcite{He_2025} found that both iterative human prompting and automated optimization within the DSPy framework~\cite{khattab2023dspycompilingdeclarativelanguage} were unreliable when gold-standard labels were scarce. Critically, none of these comparisons evaluate domain-expert prompts developed by professional practitioners, nor do they address specialized linguistic tasks where prompt quality has direct production consequences.

\paragraph{LLMs for Linguistic Tasks.}
LLMs have progressed rapidly on machine translation: early evaluations showed competitive performance on high-resource European languages but gaps in low-resource settings~\cite{jiao2023chatgptgoodtranslatoryes,hendy2023goodgptmodelsmachine}, with GPT-4 surpassing supervised baselines such as NLLB~\cite{nllbteam2022languageleftbehindscaling} in many directions while still lagging behind commercial systems in low-resource languages~\cite{zhu-etal-2024-multilingual}. More recent benchmarks show LLMs now dominate, with Gemini~2.5 Pro placing in the top evaluation cluster for 14 of 16 language pairs~\cite{kocmi-etal-2025-findings}. Fine-grained human evaluations suggest LLM translations remain comparable to junior and mid-level professional translators but fall short of senior professionals~\cite{yan2024benchmarkinggpt4humantranslators}.

For quality assessment, neural metrics such as COMET~\cite{rei-etal-2020-comet} and BLEURT~\cite{sellam-etal-2020-bleurt} correlate well with human assessment~\cite{freitag-etal-2021-experts}, and xCOMET~\cite{guerreiro2023xcomettransparentmachinetranslation} extends this with error span detection, but these learned metrics lack the explainability of prompted LLM evaluators. Kocmi and Federmann~\shortcite{kocmi2023largelanguagemodelsstateoftheart} showed that GPT-based models achieved state-of-the-art system-level correlation with human judgments, and their GEMBA-MQM metric~\cite{kocmi-federmann-2023-gemba} extended this to fine-grained error span detection. However, Fernandes et al.~\shortcite{fernandes-etal-2023-devil} and Huang et al.~\shortcite{huang-etal-2024-lost} found that predicted error spans do not align well with human annotations at the segment level, motivating refinements such as filtering LLM-predicted errors through automatic post-editing~\cite{lu-etal-2025-mqm}.

For terminology enforcement, treating terminology insertion as a post-translation step is closest to our setting. Moslem et al.~\shortcite{moslem-etal-2023-domain} showed that LLM-based post-editing can approximately double successful terminology insertion rates, while Kim et al.~\shortcite{kim-etal-2024-efficient} proposed a training-based approach using Trie-tree term extraction for specialized domains. 

For all three tasks, prompt quality is known to matter, yet no prior work has compared automatically optimized prompts against domain-expert prompts for specialized linguistic workflows.

\section{Methodology}

\subsection{Shared Experimental Setup}\label{sec:shared}

Across all tasks, we compare three prompting strategies: (a) hand-crafted zero-shot expert prompts; (b) base DSPy signatures without optimization; and (c) DSPy signatures optimized with GEPA \cite{agrawal2025gepareflectivepromptevolution}. Manual and base DSPy configurations use language-agnostic prompts across all locales, with the target locale provided as an input variable. GEPA-optimized systems also use language-agnostic prompts, except for LQA, where optimization runs independently per language pair, producing locale-specific prompts.
For the terminology insertion and LQA tasks, we compare unified single-stage systems against task-decomposed multi-stage pipelines, and evaluate all configurations with various models. All DSPy configurations interact with the LLMs through the \texttt{dspy.Predict} module.

\paragraph{Models.} We evaluate five model configurations spanning four providers: (1) GPT-4.1-mini with GPT-4.1 for reflection; (2) GPT-5.4-mini with GPT-5.4 for reflection; (3) Gemini 3.1 Flash-Lite with Gemini 3.1 Pro for reflection; (4) Claude Sonnet 4.6 with Claude Opus 4.6 for reflection; and (5) Qwen3:8B-q4\_K\_M with Qwen3:14B-q4\_K\_M for reflection, hosted locally via Ollama. All configurations use temperature 0 and max 4K tokens for execution, and temperature 1 and 32K tokens for reflection.

\paragraph{Data.} For all tasks, we use proprietary datasets spanning various domains, with gold-standard annotations produced by professional contract linguists. Annotations are reviewed internally, with in-house feedback used to refine and correct them before inclusion in the dataset. Data is split into train (20\%), validation (40\%), and test (40\%) sets, as in Agrawal et al.~\shortcite{agrawal2025gepareflectivepromptevolution}, who
allocate the majority of data to validation and test since GEPA's
reflective optimization requires relatively few training examples. Test set sizes are 1085 examples for terminology insertion, 989 for translation, and 1250 for LQA. Per-language results are omitted for space reasons but can be found in the accompanying GitHub repository. 

\paragraph{Optimization.} For GEPA optimization, we use ``light''
mode, which minimizes the number of programs proposed
and evaluated per run. We choose this setting for two reasons: it reflects the practical constraint of optimizing across three tasks, five models, and multiple objectives within a manageable compute budget; and it establishes a conservative lower bound on optimization performance that provides a fairer comparison against zero-shot baselines. Textual feedback comparing gold vs.\ predicted outputs is provided
to the optimizer in all configurations, with task-specific optimization
objectives described in each task's subsection.

\subsection{Terminology Insertion}

Given a source text, its machine translation, and relevant glossary terms, our terminology insertion system performs automatic post-editing to enforce glossary compliance in
translations where required terms are missing. If glossary violations are found, the system produces a mapping of incorrect terms in the original translation to the source terms in the glossary that is used for glossary lookup as a reasoning reference.

We experiment with terminology insertion in translations from English to six target locales (Arabic, Spanish, German, Russian, Traditional Chinese, Swedish). We follow the shared experimental setup (Section~\ref{sec:shared}), with two
task-specific details: task decomposition compares a unified glossary
term insertion system against a three-stage pipeline of violation
identification, correction, and validation; and GEPA optimization
targets three distinct objectives, detailed below.

Each example consists of a source text, original translation with approximately 70\% missing glossary terms, source and target locale codes, and a dictionary of glossary terms previously automatically determined, via stemming, as present in the source text, but missing from the machine translated text. The relevance of glossary terms varies.\footnote{Part-of-speech is not included, glossaries may contain duplications, variants of target terms could be mapped to the same source term, and capitalization of terms is not always consistent.} For example, the source text \textit{``Start with lexical search"} contains \textit{``search"} as a noun, but the detected glossary term for de-DE is a verb: \texttt{[{'sourceTerm': 'search', 'targetTerm': 'suchen'}]}. Although not relevant in context, this glossary is still given to the prompt. The manual prompts and DSPY signatures account for this uncertainty, instructing the model to disregard glossary entries that are incorrect or irrelevant and delegating the filtering decision to the LLM.

\paragraph{Optimization objectives.} We optimize for three metrics: (1) BLEU with textual feedback comparing gold vs.\ predicted glossary-integrated translations and detailing missing or incorrectly inserted glossary terms by comparing stems; (2) HTER with textual feedback; (3) LLM-as-a-judge, using the reflection model, scoring glossary-integrated translations based on correct glossary usage, fluency, no extra edits, and correct capitalization.

\paragraph{Evaluation.} We report BLEU, HTER (both using the SacreBLEU implementation) as discussed in Post~\shortcite{post2018clarityreportingbleuscores}, measuring translation quality and edit distance from gold human translations, respectively, on the predicted glossary-integrated translation. We additionally report TMR (term match rate) as the rate at which glossary term stems that are present in the gold translation appear in the predicted translation. Both unified and task-decomposed systems are judged on the final produced glossary-compliant translation.

\subsection{Translation}

Given the source locale, target locale, source text, glossary, similar translated example text, and a translation style guide, our translation system produces output text in the target language that adheres to the stylistic constraints.

Our translation dataset spans ten target locales (German, Spanish, French, Italian, Japanese, Korean, Polish, Russian, Turkish, and Chinese), each translating from English.

Each example in the test dataset contains a source locale, a target locale, and a source text. Data points may contain any combination of glossary, similar translated text, and style guide, or none at all. This mimics the variety of linguistic resources available in real-world translation settings.

\paragraph{Optimization objectives.} We optimize for the average between HTER and chrF3, as defined by Popović et al \shortcite{Popovic2015chrFCN}. We use the SacreBLEU implementation for both scores. Together, these metrics provide a strong comparison between the machine-translated text and human-translated reference. To normalize both scores to the same scale, we use the formula $score=\frac{100 - HTER}{200} + \frac{chrF3}{100}$

\paragraph{Evaluation.} We report BLEU, HTER, and chrF3 scores calculated between the LLM output text and the gold-standard reference text. The mean and median scores are produced for the micro- and macro-average across the locales.

\subsection{Language Quality Assessment}

Given a source sentence and its machine translation, our LQA system predicts whether a translation error exists and, if so, its type and severity using an MQM-compliant schema of 15 error types and 3 severity levels.\footnote{Error types: Mistranslation, Omission, Addition, Grammar, Spelling, Punctuation, Unidiomatic, Register, Inconsistency, Language Variety, Whitespace, Markup/Technical, Locale Formatting, Duplication, Culture-specific reference; Severities: Minor, Major, Critical}

We investigate the LQA task for translations from English into six target languages (Arabic, Spanish, German, Russian, Simplified Chinese, Swedish). Task decomposition compares unified error detection and characterization against a two-stage detection-then-characterization pipeline. 

Each data instance consists of source text, target translation, and locale code as inputs, with binary error presence and a JSON array of errors, specifying category, severity, and annotator comment, as labels. Multiple errors per instance are supported. Error-bearing examples
naturally constitute the majority of the data ($\sim$65\% across training, validation, and test splits).

\paragraph{Optimization objectives.} Single-stage systems optimize a composite metric: 0.0 for detection mismatches, 1.0 for correct clean predictions, and $0.5 + 0.25 \times \text{Category F1} + 0.25 \times \text{Severity F1}$ for true positives. Given the class distribution, the optimizer cannot gain by defaulting
to predicting no error at the example level. Two-stage systems optimize detection with a binary accuracy metric, while characterization (applied only to error-bearing examples) optimizes a hybrid score: $0.5 \times \text{Category F1} + 0.5 \times \text{Severity F1}$.

\paragraph{Evaluation.} We report four primary metrics: Detection F1, Category F1, Severity F1, and Pearson's correlation with gold MQM scores.\footnote{MQM is calculated by weighting errors by severity (Minor=1, Major=5, Critical=25), normalizing by sentence length, and subtracting from a perfect base score of 100.} All metrics are computed end-to-end on system outputs. Cross-locale summary statistics are macro-averaged, weighting each locale equally regardless of test set size.

\section{Results}
\subsection{Terminology Insertion}

Table~\ref{tab:gtioverallresults} presents the results across all configurations. We highlight the key findings below.

\paragraph{Manual and optimized prompts achieve largely comparable terminology insertion quality across models.} Manual prompts achieve BLEU scores higher than or equal to base DSPy for most models, with Qwen3:8B-q4\_K\_M as an exception where Base DSPy matches or exceeds Manual. Compared to optimized DSPy configurations, Manual BLEU scores are slightly higher or equivalent (1--2 point differences), though for Claude Sonnet 4.6 and GPT-4.1-mini decomposed, optimized prompts match or marginally surpass Manual. HTER differences are slightly more pronounced but remain non statistically significant: unified manual prompts achieve lower HTER than optimized DSPy for three of five models (GPT-4.1-mini, GPT-5.4-mini, and Gemini 3.1 Flash-Lite), with gaps of 2--4 points, while Claude Sonnet 4.6 shows a marginally lower HTER for optimized DSPy and Qwen3:8B-q4\_K\_M favors DSPy configurations more broadly. Optimized prompts achieved the highest term match rates, with statistically significant gains for half of the models, reflecting that optimization prioritizes glossary term presence over fluency as measured by HTER or BLEU.

\paragraph{Optimization objective has limited impact for most models, with notable exceptions.} Across three of five models (GPT-4.1-mini, Gemini 3.1 Flash-Lite, and Qwen3:8B-q4\_K\_M), the three optimization
objectives produced near-identical results. For GPT-5.4-mini and
Claude Sonnet 4.6, the choice of objective did affect BLEU and HTER,
with LLM-as-a-judge yielding the best scores. However, metric-aligned
optimization does not guarantee metric-specific gains: optimizing
against BLEU did not consistently produce the highest BLEU scores, nor
did optimizing against HTER consistently minimize edit distance. In
all cases, differences across objectives within a model were smaller
than differences across models, with Gemini 3.1 Flash-Lite performing
best overall regardless of optimization objective.

\begin{table*}[t]
  \centering
  \small
  {
\setlength{\tabcolsep}{9pt}
\begin{tabular}{llcccccc}
\hline
& & \multicolumn{2}{c}{\textbf{BLEU}}
& \multicolumn{2}{c}{\textbf{HTER}}
& \multicolumn{2}{c}{\textbf{TMR}} \\
\cline{3-8}
\textbf{Model} & \textbf{Prompt} 
& \textbf{Uni} & \textbf{Dec}
& \textbf{Uni} & \textbf{Dec}
& \textbf{Uni} & \textbf{Dec} \\
\hline
GPT-4.1-mini 
& Manual 
& \textbf{0.48} & 0.46 & \textbf{0.53} & \textbf{0.56} & 0.86 & 0.77 \\
& Base 
& 0.45 & 0.45 & 0.55 & 0.57 & 0.87 & 0.81 \\
& Opt.\ DSPy (best)
& 0.46\textsuperscript{(2,3)} & \textbf{0.49}\textsuperscript{(1)} 
& 0.55\textsuperscript{(3)} & \textbf{0.56}\textsuperscript{(1,3)} & \textbf{0.88}\textsuperscript{(1)} & \textbf{0.91}\textsuperscript{(2)}$^\ddagger$ \\
\cline{1-8}

GPT-5.4-mini 
& Manual 
& \textbf{0.50} & \textbf{0.48} & \textbf{0.50} & \textbf{0.55} & 0.90 & 0.82 \\
& Base DSPy
& 0.45 & 0.45 & 0.56 & 0.57 & 0.85 & 0.78 \\
& Opt.\ DSPy (best)
& 0.49\textsuperscript{(3)} & 0.47\textsuperscript{(1,2)}
& 0.52\textsuperscript{(3)} & \textbf{0.55}\textsuperscript{(2)} 
& \textbf{0.91}\textsuperscript{(3)} & \textbf{0.91}\textsuperscript{(3)}$^\ddagger$ \\
\cline{1-8}

Gemini 3.1 Flash-Lite
& Manual 
& \textbf{0.52} & \textbf{0.51} & \textbf{0.47} & \textbf{0.50} & \textbf{0.94} & 0.92 \\
& Base DSPy
& 0.50 & 0.50 & 0.51 & \textbf{0.50} & 0.92 & 0.91 \\
& Opt.\ DSPy (best)
& 0.50\textsuperscript{(1,2,3)} & \textbf{0.51}\textsuperscript{(3)}
& 0.51\textsuperscript{(1,2,3)} & 0.52\textsuperscript{(2,3)}
& 0.92\textsuperscript{(1,2,3)} & \textbf{0.94}\textsuperscript{(2)} \\
\cline{1-8}
Claude Sonnet 4.6
& Manual 
& 0.49 & 0.47 & 0.51 & 0.55 & 0.90 & 0.84 \\
& Base DSPy 
& 0.48 & 0.47 & 0.54 & 0.55 & \textbf{0.93} & \textbf{0.93} \\
& Opt.\ DSPy (best)
& \textbf{0.50}\textsuperscript{(3)} & \textbf{0.50}\textsuperscript{(3)}
& \textbf{0.50}\textsuperscript{(3)} & \textbf{0.53}\textsuperscript{(3)} 
& \textbf{0.93}\textsuperscript{(3)} & \textbf{0.93}\textsuperscript{(3)}\\
\cline{1-8}
Qwen3:8B-q4\_K\_M
& Manual 
& 0.38 & 0.38 & 0.63 & 0.63 & 0.74 & 0.74 \\
& Base DSPy 
& 0.38 & \textbf{0.40} & \textbf{0.59} & \textbf{0.61} & 0.74 & 0.76 \\
& Opt.\ DSPy (best)
& 0.38\textsuperscript{(1,2,3)} & \textbf{0.40}\textsuperscript{(2)}
& \textbf{0.59}\textsuperscript{(1,2,3)} & 0.62\textsuperscript{(1,2)} 
& 0.74\textsuperscript{(1,2,3)} & \textbf{0.82}\textsuperscript{(1)}$^\ddagger$\\
\cline{1-8}
\hline
\end{tabular}
    }
  \caption{Summary performance for the Terminology Insertion task, macro-averaged across six target locales (ar, de-DE, es-ES, ru-RU, sv-SE, zh-TW). Uni = unified (single-stage);
Dec = decomposed (three-stage) N: 1085 (ar=85, de-DE=200, es-ES=200, ru-RU=200, sv-SE=200, zh-TW=200). Each Opt.\ DSPy cell reports the score from the
best-performing optimization objective for that metric and
architecture; superscripts identify which:\textsuperscript{(1)}~BLEU, \textsuperscript{(2)}~HTER,
\textsuperscript{(3)}~LLM-based judge. Best
value per model and column in \textbf{bold}. $^\ddagger$~$p<0.01$ vs.\ the next-best configuration (paired stratified bootstrap, 1{,}000 iterations, Holm--Bonferroni corrected per model).}
  \label{tab:gtioverallresults}
\end{table*}

\subsection{Translation}

Table~\ref{tab:translation_summary} presents the  results for the translation task with the three metrics BLEU, HTER, and ChrF3. We highlight the key findings below.

\paragraph{Manual prompts generally win for newer models.} On an older model like OpenAI's GPT-4.1-mini, DSPy prompts consistently win on all metrics. However, some modern models (GPT-5.4-mini and Gemini 3.1 Flash-Lite) actually perform best with manually crafted prompts. Claude Sonnet 4.6 is an exception.

\paragraph{Alternative prompting strategies perform statistically close to the best strategy.} Although one strategy tends to dominate for a given model, it is typically not significantly better than the second-best alternative.
Gemini 3.1 Flash-Lite is an exception: the manual prompt significantly outperforms Opt.~DSPy on BLEU mean ($p<0.01$).
For all other models, alternative strategies can be substituted with minimal performance loss.

\paragraph{The open-weight model performed significantly worse than any proprietary model on translation.} For Qwen3:8B-q4\_K\_M, generated prompts provide a statistically confirmed benefit: Base~DSPy significantly outperforms the manual prompt on BLEU mean. However, even with this improvement, the model falls far short of every proprietary model across all metrics and strategies.

\paragraph{The best prompting strategy for each model is generally consistent across scores.} The prompts were optimized against a mixture of HTER and ChrF3, but the best BLEU scores also coincide with the best overall prompt. This lends cross-metric confidence to the quality of the generated translations.

\begin{table*}
  \centering
  \small
  \setlength{\tabcolsep}{12pt}
  \begin{tabular}{llcccccc}
    \hline
    & & \multicolumn{2}{c}{\textbf{BLEU}} & \multicolumn{2}{c}{\textbf{HTER}} & \multicolumn{2}{c}{\textbf{ChrF3}} \\
    \cline{3-4} \cline{5-6} \cline{7-8}
    \textbf{Model} & \textbf{Prompt} & \textbf{Mean} & \textbf{Median} & \textbf{Mean} & \textbf{Median} & \textbf{Mean} & \textbf{Median} \\
    \hline
    GPT-4.1-mini & Manual & 57.93 & 58.77 & 30.32 & 25.73 & 70.55 & 70.91 \\
     & Base DSPy & 58.21 & \textbf{60.15} & 30.27 & \textbf{25.41} & 70.59 & 71.78 \\
     & Opt. DSPy & \textbf{59.33} & 60.13 & \textbf{29.21} & 25.79 & \textbf{71.27} & \textbf{72.64} \\
    \hline
    GPT-5.4-mini & Manual & \textbf{57.63} & \textbf{57.82} & \textbf{31.80} & \textbf{28.03} & \textbf{69.69} & \textbf{69.79} \\
     & Base DSPy & 55.60 & 55.19 & 34.07 & 29.55 & 68.88 & 69.06 \\
     & Opt. DSPy & 56.16 & 55.81 & 33.16 & 29.26 & 69.18 & 69.41 \\
    \hline
    Gemini 3.1 Flash-Lite & Manual & \textbf{60.75}$^{\ddagger}$ & \textbf{61.31} & \textbf{29.20} & \textbf{25.04} & \textbf{72.61} & \textbf{73.45} \\
     & Base DSPy & 58.43 & 58.51 & 30.53 & 26.80 & 71.16 & 72.20 \\
     & Opt. DSPy & 58.53 & 59.40 & 30.00 & 25.60 & 71.47 & 72.57 \\
    \hline
    Claude Sonnet 4.6 & Manual & 58.81 & 59.32 & 29.49 & 25.21 & 71.54 & 72.97 \\
     & Base DSPy & 58.42 & 58.99 & 30.31 & 25.80 & 71.23 & 72.24 \\
     & Opt. DSPy & \textbf{59.13} & \textbf{60.97} & \textbf{29.47} & \textbf{24.65} & \textbf{71.86} & \textbf{73.53} \\
    \hline
    Qwen3:8B-q4\_K\_M & Manual & 46.06 & 42.74 & 40.41 & 39.16 & 61.53 & 60.64 \\
     & Base DSPy & \textbf{48.03} & 44.37 & \textbf{38.88} & 37.14 & \textbf{62.18} & 61.29 \\
     & Opt. DSPy & 47.33 & \textbf{44.81} & 39.44 & \textbf{36.92} & 61.98 & \textbf{61.76} \\
    \hline
  \end{tabular}

  \caption{Summary performance for the Translation task, macro-averaged translation scores across ten target locales. N: 972 (de-DE=140, es-ES=69, fr-FR=133, it-IT=15, ja-JP=164, ko-KR=170, pl-PL=60, ru-RU=50, tr-TR=44, zh-CN=127). Mean and median of BLEU, HTER, and ChrF3. BLEU, ChrF3 = higher is better, HTER = lower is better. Best value per model and column in \textbf{bold}. $^\dagger$~$p<0.05$, $^\ddagger$~$p<0.01$ vs.\ the next-best configuration (paired stratified bootstrap, 1{,}000 iterations, Holm--Bonferroni corrected per model).}
  \label{tab:translation_summary}
\end{table*}

\begin{table*}[t]
  \centering
  \small
  \setlength{\tabcolsep}{9pt}
  \begin{tabular}{llcccccccc}
    \hline
    & & \multicolumn{2}{c}{\textbf{Det F1}} & \multicolumn{2}{c}{\textbf{Cat F1}} & \multicolumn{2}{c}{\textbf{Sev F1}} & \multicolumn{2}{c}{\textbf{MQM $r$}} \\
    \cline{3-4} \cline{5-6} \cline{7-8} \cline{9-10}
    \textbf{Model} & \textbf{Prompt} & \textbf{Uni} & \textbf{Dec} & \textbf{Uni} & \textbf{Dec} & \textbf{Uni} & \textbf{Dec} & \textbf{Uni} & \textbf{Dec} \\
    \hline
    GPT-4.1-mini & Manual & \textbf{0.64} & 0.54 & 0.36 & 0.39 & \textbf{0.54}$^\ddagger$ & \textbf{0.53}$^\ddagger$ & 0.15 & 0.18 \\
     & Base DSPy & 0.52 & 0.43 & 0.39 & \textbf{0.44} & 0.48 & 0.52 & \textbf{0.21} & \textbf{0.18} \\
     & Opt. DSPy & 0.63 & \textbf{0.60}$^\ddagger$ & \textbf{0.41} & 0.40 & 0.46 & 0.50 & 0.20 & 0.16 \\
    \hline
    GPT-5.4-mini & Manual & \textbf{0.60}$^\ddagger$ & 0.48 & 0.36 & 0.36 & 0.45 & 0.38 & 0.12 & 0.08 \\
     & Base DSPy & 0.53 & 0.41 & 0.33 & 0.36 & 0.45 & 0.33 & \textbf{0.15} & 0.11 \\
     & Opt. DSPy & 0.49 & \textbf{0.55}$^\ddagger$ & \textbf{0.49}$^\dagger$ & \textbf{0.42}$^\ddagger$ & \textbf{0.52} & \textbf{0.48}$^\ddagger$ & 0.12 & \textbf{0.14} \\
    \hline
    Gemini 3.1 Flash-Lite & Manual & \textbf{0.64}$^\ddagger$ & 0.42 & 0.41 & 0.42 & 0.61 & 0.51 & \textbf{0.23} & 0.09 \\
     & Base DSPy & 0.51 & 0.34 & 0.41 & 0.44 & 0.55 & 0.54 & 0.16 & 0.17 \\
     & Opt. DSPy & 0.56 & \textbf{0.59}$^\ddagger$ & \textbf{0.51}$^\ddagger$ & \textbf{0.53}$^\ddagger$ & \textbf{0.68}$^\ddagger$ & \textbf{0.65}$^\ddagger$ & 0.21 & \textbf{0.25}$^\dagger$ \\
    \hline
    Claude Sonnet 4.6 & Manual & \textbf{0.56} & 0.43 & 0.43 & 0.46 & 0.61 & 0.51 & \textbf{0.11} & 0.08 \\
     & Base DSPy & 0.46 & 0.37 & 0.43 & 0.48 & \textbf{0.63} & 0.52 & 0.10 & \textbf{0.12} \\
     & Opt. DSPy & 0.53 & \textbf{0.53}$^\ddagger$ & \textbf{0.48} & \textbf{0.53}$^\ddagger$ & 0.59 & \textbf{0.61}$^\ddagger$ & 0.08 & 0.10 \\
    \hline
    Qwen3:8B-q4\_K\_M & Manual & \textbf{0.60}$^\ddagger$ & \textbf{0.54}  & \textbf{0.32} & 0.28 & 0.44 & 0.39 & \textbf{0.19} & 0.16 \\
     & Base DSPy & 0.53 & 0.41 & 0.31 & 0.30 & 0.38 & 0.30 & 0.16 & 0.13 \\
     & Opt. DSPy & 0.51 & 0.48 & 0.31 & \textbf{0.31} & \textbf{0.55}$^\ddagger$ & \textbf{0.45}$^\ddagger$ & 0.15 & \textbf{0.18} \\
    \hline
  \end{tabular}
  
  \caption{Summary performance for the LQA task, macro-averaged across six target locales. Uni = unified (single-stage); Dec = decomposed (two-stage). N: 1250 (ar=91, de-DE=353, es-ES=362, ru-RU=80, sv-SE=245, zh-CN=119). Best value per model and column in \textbf{bold}. $^\dagger$~$p<0.05$, $^\ddagger$~$p<0.01$ vs.\ the next-best configuration (paired stratified bootstrap, 1{,}000 iterations, Holm--Bonferroni corrected per model).}
  \label{tab:autolqa}
\end{table*}

\subsection{Language Quality Assessment}

Table~\ref{tab:autolqa} presents the results across all configurations. We highlight the key findings below.

\paragraph{GEPA optimization consistently improves base DSPy signatures across model families.} In all models, GEPA-optimized unified prompts match or exceed manual prompt performance on at least one primary metric. The pattern is most pronounced for characterization: decomposed optimized prompts achieve the highest Category F1 in four of five models, with Severity F1 showing a similar pattern.

\paragraph{Manual expert prompts retain an advantage on error detection.} Across all five models, manual unified prompts achieve the highest Detection F1. However, this advantage over optimized prompts is statistically significant in only three of five models. In decomposed systems, optimized prompts significantly outperform manual prompts on Detection F1 in all models except Qwen3:8B-q4\_K\_M. 

\paragraph{Unified systems outperform decomposed pipelines across all models.} This finding generalizes robustly: in all five models, unified systems outperform their two-stage counterparts on Detection F1, typically by 0.05–0.20 points. We note that our significance tests compare prompt strategies within each architecture and do not directly test the unified-versus-decomposed comparison.

\section{Conclusions}
 
We presented, to our knowledge, the first systematic comparison of expert-crafted prompts against automatically optimized prompts for specialized linguistic tasks, evaluating across terminology insertion, translation, and LQA with five model configurations from four providers.
 
\paragraph{Central findings.} GEPA optimization elevates minimal DSPy signatures, often closing the gap to expert-level performance, but the degree and direction of remaining differences are task-dependent. In terminology insertion, expert and optimized prompts are largely statistically indistinguishable on translation quality (BLEU, HTER), with optimization producing significantly higher term match rates for several models but not consistently improving fluency. In translation, most model–prompt differences are not statistically significant, with the exception of Gemini 3.1 Flash-Lite, where expert prompts significantly outperform optimized ones on BLEU. In LQA, optimization yields statistically significant gains over manual prompts on error characterization for several models, more consistently for Gemini 3.1 Flash-Lite. The degree to which optimization improves over base DSPy signatures also varies by task: gains are largest and most consistent in translation and LQA, while terminology insertion shows smaller improvements. The magnitude of optimization gains also varies by model: Qwen3:8B-q4\_K\_M, while performing worse than proprietary models on translation and terminology insertion, narrowed the gap on several LQA metrics, with optimization over base DSPy producing the largest single gain observed across all tasks and models, suggesting optimization may yield larger returns over base DSPy when the model is weaker.

\paragraph{Practical implications.} We highlight three concrete takeaways for practitioners. First, unified single-stage systems generally match or outperform decomposed multi-stage pipelines---most clearly for error detection in LQA (0.05--0.20 points on Detection F1) and edit distance in terminology insertion---suggesting end-to-end designs should be preferred unless there is a specific reason to decompose. Second, GEPA reliably elevates minimal DSPy signatures across tasks and models, making it a strong default when labeled data is available. Third, within unified systems, expert prompts retain an advantage for LQA error detection while optimization helps characterization ---no single approach wins on every sub-metric. This suggests that combining the two, for example by seeding GEPA with expert prompts and letting it refine instructions, may be especially valuable for LQA. Practitioners should choose between approaches, or combine them, based on their available resources (labeled data, human effort, compute) and task requirements, rather than treating one as inherently superior.

\paragraph{Limitations.} This study has several limitations. First, the comparison involves an inherent asymmetry in how each approach uses labeled data. GEPA optimization programmatically exploits gold-standard training and validation splits through automated feedback loops, whereas expert prompt engineering draws on domain knowledge, iterative testing, and, where available, selective inspection of examples. Second, the two approaches are not mutually exclusive: a domain expert could use GEPA optimization as a starting point and manually refine the resulting prompts, or conversely, use expert-crafted prompts as seed inputs for automatic optimization. Our study treats them as independent conditions, but a combined workflow may outperform either in isolation. Third, several design choices were scoped for practical reasons: we evaluate a single optimizer (GEPA) in its ``light" mode with the \texttt{dspy.Predict} module, we do not evaluate few-shot or chain-of-thought manual prompts, and all evaluation is automatic without human judgments. Each of these choices may underestimate the ceiling of either approach: stronger optimizers, richer expert baselines, and human evaluation could all shift the results, but they were necessary to keep the study tractable across three tasks, five models, and multiple conditions. Fourth, all language pairs are English-centric, reflecting the availability of gold-standard data per task; results may not generalize to non-English-centric directions.

\paragraph{Future work.} Future work could address these gaps by comparing additional optimizers and optimization modes, evaluating few-shot expert prompts, evaluating hybrid workflows that combine expert knowledge with automatic optimization, and conducting a cost-benefit analysis of manual versus automatic prompt development across the product lifecycle.

\section{Sustainability Statement}
This study involves no model training or fine-tuning; all experiments consist of inference-time prompt optimization and evaluation. Four of five model configurations rely on commercial APIs (OpenAI, Google, Anthropic), for which precise energy consumption cannot be estimated. The fifth uses locally hosted open-weight models run via Ollama on consumer-grade laptops. GEPA's ``light" optimization mode minimizes candidate programs per run, keeping the overall computational footprint modest. We acknowledge that repeated API calls across multiple tasks, models, and optimization objectives contribute to cumulative energy use that cannot be precisely quantified for API-based configurations.

\bibliography{eamt26}
\bibliographystyle{eamt26}

\onecolumn

\end{document}